\documentclass[12pt, reqno]{amsart}

\author[M. Caprio, Y. Sale, E. Hüllermeier]{Michele Caprio \qquad Yusuf Sale \qquad Eyke Hüllermeier}
\address{The University of Manchester, Oxford Road, Manchester, UK M13 9PL}
\email{michele.caprio@manchester.ac.uk}
\address{Ludwig-Maximilian University, Akademiestraße 7, Munich, Germany 80799}
\email{\{yusuf.sale,eyke\}@ifi.lmu.de}

\title{Conformal Prediction Regions are Imprecise Highest Density Regions}

\usepackage[T1]{fontenc}
\usepackage{amsmath}
\usepackage{amssymb}
\usepackage{lscape}
\usepackage{amsthm}
\usepackage[left=2.5cm, right=2.5cm, top=3cm]{geometry}
\usepackage{hyperref}
\usepackage{algorithm}
\usepackage{bbm}
\usepackage{booktabs}
\usepackage{algorithm}
\usepackage{algpseudocode}
\usepackage{tikz}
\usepackage{caption}
\usepackage{subcaption}
\usepackage{fancyhdr}
\usepackage[inline]{enumitem}
\usepackage{comment}
\usepackage{nicefrac}
\usepackage{bm}
\usepackage{mathrsfs}
\usepackage{multirow}
\usepackage{graphicx}
\usepackage[utf8]{inputenc}
\usepackage{cancel}
\usepackage{mathtools}
\usepackage{natbib}

\newcommand{\vertiii}[1]{{\left\vert\kern-0.25ex\left\vert\kern-0.25ex\left\vert #1 
    \right\vert\kern-0.25ex\right\vert\kern-0.25ex\right\vert}}

\AtBeginDocument{%
   \def\MR#1{}
}

\makeatletter
\def\algbackskip{\hskip-\ALG@thistlm}
\makeatother

\newtheorem{thm}{Theorem}[section]

\theoremstyle{definition} 
\newtheorem{defi}[thm]{Definition}
\let\olddefi\defi
\renewcommand{\defi}{\olddefi\normalfont}
\newtheorem{rmk}{Remark}
\let\oldrmk\rmk
\renewcommand{\rmk}{\oldrmk\normalfont}

\DeclareMathOperator*{\argmax}{arg\,max}

\newtheorem{theorem}{Theorem}
\newtheorem{lemma}[theorem]{Lemma}
\newtheorem{proposition}[theorem]{Proposition}
\newtheorem{corollary}{Corollary}[theorem]

\newtheorem{definition}[theorem]{Definition}
\newtheorem{remark}[theorem]{Remark}

\hypersetup{
    pdftitle={prove},
    pdfauthor={autori},
    pdfmenubar=false,
    pdffitwindow=true,
    pdfstartview=FitH,
    colorlinks=true,
    linkcolor=blue,
    citecolor=green,
    urlcolor=cyan
}

\providecommand{\MR}[1]{}

\providecommand{\MR}{\relax\ifhmode\unskip\space\fi MR }

\begin{document}

\begin{abstract}
Recently, Cella and Martin proved how, under an assumption called {\em consonance}, a credal set (i.e. a closed and convex set of probabilities) can be derived from the conformal transducer associated with transductive conformal prediction. We show that the Imprecise Highest Density Region (IHDR) associated with such a credal set corresponds to the classical Conformal Prediction Region. In proving this result, we establish a new relationship between Conformal Prediction and Imprecise Probability (IP) theories, via the IP concept of a cloud. A byproduct of our presentation is the discovery that consonant plausibility functions are monoid homomorphisms, a new algebraic property of an IP tool.
\end{abstract}

\maketitle
\thispagestyle{empty}

\section{Introduction}\label{sec:intro}
Conformal prediction (CP) is a methodology introduced by \citet{vovk2005algorithmic} whose main goal is to output a prediction for the next observation's value, given a collection of data points, which takes into account the uncertainty faced by the agent. Loosely, CP accounts for the latter by producing predictions in the form of \textit{conformal prediction regions} (CPRs), subsets of the output space that are likely to cover the target of prediction -- the true outcome -- with high probability, for all possible exchangeable probability distributions on the output space. CP is a model-free approach: it does not require any distributional assumptions on the underlying data-generating process, making it a versatile and widely applicable tool e.g. for statistical inference. 

A key part of (transductive) CP is to derive the conformal transducer $\pi$, a ``score'' between $0$ and $1$ that measures how ``conformal'' an element $y$ of the output space $\mathbb{Y}$ is to the data previously available to the user. The ``more conformal'' element $y$ is, the closer its ``score'' is to $1$. In their work \citep{cella2022validity,cella2021valid}, Cella and Martin show that, once $\pi$ is available, an upper probability $\overline{\Pi}$ (that can be thought of as the upper envelope of a closed and convex set of probabilities, i.e. a credal set) can be derived from it. This is possible under a given assumption called \textit{consonance}, which, roughly, tells us that there is at least an element $y\in\mathbb Y$ such that $\pi(y)=1$.\footnote{Consonance can for instance be obtained ``artificially'' by putting to $1$ the value of $\pi$ for $\argmax_{y} \pi(y)$ \citep[Section 7]{cella2021valid}.} Under consonance, upper probability $\overline{\Pi}$ is defined as $\overline{\Pi}(A)=\sup_{y\in A} \pi(y)$, for all $A \subseteq \mathbb Y$, and the corresponding credal set is given by $\mathcal{M}(\overline{\Pi})=\{P : P(A) \leq \overline{\Pi}(A) \text{, } \forall A \subseteq \mathbb Y\}$. That is, $\mathcal{M}(\overline{\Pi})$ contains all the probabilities on $\mathbb{Y}$ that are set-wise dominated by $\overline{\Pi}$.

\begin{figure*}[h!]
\centering
\includegraphics[width=.6\textwidth]{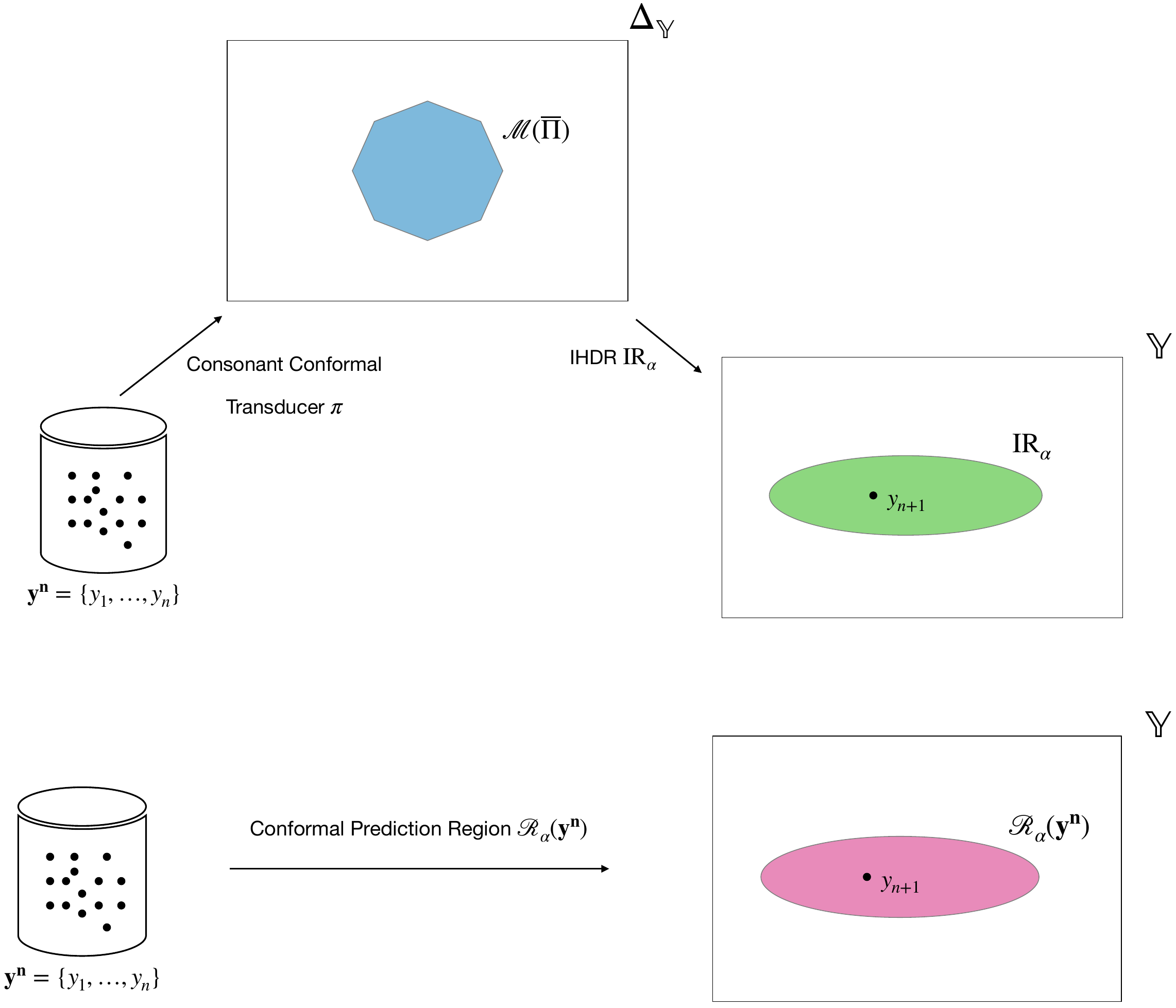}
\caption{Top: Our proposed, ``indirect'' methodology to derive a prediction region. We first use the consonant conformal transducer $\pi$ to derive credal set $\mathcal{M}(\overline{\Pi})$, and then extract from the latter the IHDR $\text{IR}_\alpha$. Bottom: Classical CP methodology, in which the Conformal Prediction Region is obtained as in \eqref{eq_imp5}.}\label{fig1}
\end{figure*}

In this work, we further the study of the relationship between Conformal Prediction and Imprecise Probabilities (IPs). We 
show that the Imprecise Highest Density Region (IHDR) associated with $\mathcal{M}(\overline{\Pi})$ is equivalent to
the classical Conformal Prediction Region (CPR), and it retains the same (uniform) probabilistic guarantee. Loosely, an IHDR \citep{coolen1992imprecise} is the IP counterpart of a Highest Density Region, which in turn can be thought of as a Bayesian version of a Confidence Interval. It is a collection of elements of the output space $\mathbb Y$, that \textit{all} the elements of credal set $\mathcal{M}(\overline{\Pi})$ indicate as being correct with high probability $1-\alpha$, with significance level $\alpha$ chosen by the user. 
A visual representation of our proposed method is given in Figure \ref{fig1}. In proving this result, we also relate Conformal Prediction to the concept of cloud \citep{neumaier}, and discover a very interesting algebraic property of consonant plausibility functions like the upper probability $\overline{\Pi}$.

%

We conclude with a discussion on the open problems and unanswered questions on the nature of CP, and on the relationship between CP and IP theory.

The paper is organized as follows. Section \ref{c_pred} recalls the basics of (transductive) Conformal Prediction. Section \ref{ips} introduces the concepts from Imprecise Probability (IP) theory that are needed in our work. Section \ref{rel1} presents the relationship between conformal prediction and the IP notion of plausibility functions noted by \citet{cella2022validity,cella2021valid}. Section \ref{ihdr_sec} introduces Imprecise Highest Density Regions (IHDRs). 
In Section \ref{narrower}, we derive the IHDR of our conformally-built credal set, that corresponds to the classical conformal prediction region.
Section \ref{concl} concludes our work and discusses  further interesting open problems. 

\section{Preliminaries}\label{sec:prelim}
\subsection{Conformal Prediction}\label{c_pred}
This section is based on \citet[Section 4.1]{cella2022validity}, who summarize work by \citet{shafer2008tutorial} and \citet{vovk2005algorithmic} on transductive conformal prediction. 

Suppose that there is an exchangeable process $Y_1,Y_2,\ldots$ with distribution $\mathbf{P}$, where each $Y_i$ is a random element taking values in $\mathbb{Y}$. Throughout the paper, we denote by $\mathbf{P}$ the true data generating process, and by $P$ a generic element of $\Delta_{\mathbb{Y}}$, the space of all (countably additive) probability measures on $\mathbb{Y}$. Recall that a sequence is exchangeable if, for any $k\in\mathbb{N}$ and any permutation $\tau$, the two random vectors $(Y_1,\ldots,Y_k)^\top$ and  $(Y_{\tau(1)},\ldots,Y_{\tau(k)})^\top$ have the same joint distribution. This  implies that the marginal distributions of the $Y_i$’s are the same.\footnote{Exchangeability does not imply independence, so the standard i.i.d.\ setup is in a sense more restrictive. Note also that we are not assuming any parametric form for the distribution $\mathbf{P}$.} 

We want to solve the following statistical problem. Suppose we observe the first $n$ terms of the process, that is, $\mathbf{Y^n} = (Y_1, \ldots , Y_n)^\top$. With
this data, and the assumption of exchangeability, the goal is to predict $Y_{n+1}$ using a method that is valid or reliable in a certain sense. 

Let $\mathbf{Y^{n+1}}=(\mathbf{Y^n},Y_{n+1})^\top$ be an $(n+1)$-dimensional vector consisting of the observable $\mathbf{Y^n}$ and the yet-to-be-observed value $Y_{n+1}$. Consider the transform 
$$
\mathbf{Y^{n+1}} \rightarrow \mathbf{T^{n+1}}=(T_1,\ldots,T_{n+1})^\top
$$ 
defined by the rule 
$$
T_i\coloneqq \psi_i \left(\mathbf{Y^{n+1}} \right) \equiv \Psi \left(\mathbf{\mathbf{y^{n+1}_{-i}}},y_i \right) 
$$
for all $i\in\{1,\ldots,n+1\}$, 
where $\mathbf{\mathbf{y^{n+1}_{-i}}}=\mathbf{y^{n+1}}\setminus\{y_i\}$ and $\Psi:\mathbb{Y}^n \times\mathbb{Y} \rightarrow \mathbb{R}$ is a fixed function that is invariant to permutations in its first vector argument. Function $\Psi$, which is called a \textit{non-conformity measure}, is constructed in such as way that $\psi_i(\mathbf{y^{n+1}})$ is small if and only if $y_i$ agrees with -- i.e. is ``close to'' -- a prediction based on the data $\mathbf{\mathbf{y^{n+1}_{-i}}}$. Or, the other way around, large values $\psi_i(\mathbf{y^{n+1}})$ suggest that the observation $y_i$ is ``strange'' and does not conform to the rest of the data $\mathbf{\mathbf{y^{n+1}_{-i}}}$.
The key idea is to define $\psi_i(\mathbf{y^{n+1}})$ in a way that allows one to compare $y_i$ to a suitable summary of $\mathbf{\mathbf{y^{n+1}_{-i}}}$, e.g.\  $\psi_i(\mathbf{y^{n+1}})=|\text{mean}(\mathbf{\mathbf{y^{n+1}_{-i}}})-y_i|$, for all $i\in\{1,\ldots,n+1\}$. Notice that transformation $\mathbf{Y^{n+1}} \rightarrow \mathbf{T^{n+1}}$ preserves exchangeability. 
%

As the value $Y_{n+1}$ has not yet been observed, and is actually the target of prediction, the above calculations cannot be carried out exactly. Nevertheless, the exchangeability-preserving properties of the transformations described above provide a procedure to rank candidate values $\tilde{y}$ of $Y_{n+1}$ based on the observed $\mathbf{Y^n}=\mathbf{y^n}$, as shown in Algorithm \ref{algo1}.

\begin{algorithm}
\caption{Conformal prediction (CP)}\label{algo1}
\begin{algorithmic}
\State Initialize: data $\mathbf{y^n}$, non-conformity measure $\Psi$, grid of $\tilde{y}$ values
\For{each $\tilde{y}$ value in the grid} 
\State set $y_{n+1}=\tilde{y}$ and write $\mathbf{y^{n+1}}=\mathbf{y^n} \cup \{y_{n+1}\}$;
\State define $T_i=\psi_i(\mathbf{y^{n+1}})$, for all $i\in\{1,\ldots,n+1\}$;
\State evaluate $\pi(\tilde{y},\mathbf{y^n})=(n+1)^{-1}\sum_{i=1}^{n+1} \mathbbm{1}[T_i \geq T_{n+1}]$;
\EndFor
\State return $\pi(\tilde{y},\mathbf{y^n})$ for each $\tilde{y}$ on the grid.
\end{algorithmic}
\end{algorithm}

The output of Algorithm \ref{algo1} is a data-dependent function $\tilde{y} \mapsto \pi(\tilde{y},\mathbf{y^n})$ that can be interpreted as a measure of plausibility of the assertion that $Y_{n+1} =  \tilde{y}$, given data $\mathbf{y^n}$. \citet{vovk2005algorithmic} refer to the function $\pi$ as \textit{conformal transducer}. 
Conformal transducer $\pi$ plays a key role in the construction of \textit{conformal prediction regions} (CPRs). For any $\alpha\in [0,1]$, the $\alpha$-level CPR is defined as \citep[Equation (2)]{vovk-trans}
\begin{equation}\label{eq_imp5}
    \mathscr{R}_\alpha(\mathbf{y^n})\coloneqq\{y_{n+1}\in \mathbb{Y} : \pi(y_{n+1},\mathbf{y^n})> \alpha\},
\end{equation}
and it satisfies
\begin{equation}\label{eq_imp6}
    P\left[Y_{n+1}\in\mathscr{R}_\alpha(\mathbf{y^n})\right] \geq 1-\alpha,
\end{equation}
uniformly in $n$ and in $P$ \citep{vovk2005algorithmic}. That is, \eqref{eq_imp6} is satisfied for all $n\in\mathbb{N}$ and all exchangeable distributions $P$ on $\mathbb{Y}$. Note that \eqref{eq_imp6} holds regardless of the choice of the non-conformity measure $\Psi$. However, the choice of this function is crucial in terms of the {\em efficiency} of conformal prediction, that is, the {\em size} of the prediction regions.

Transductive conformal prediction (TCP) is not the only way of carrying out conformal prediction. There exists another type called inductive (or split) conformal prediction (ICP). It assumes exchangeability and it is used to build the same region $\mathscr{R}_\alpha(\mathbf{y^n})$ in \eqref{eq_imp5} having the same guarantee \eqref{eq_imp6}, while being computationally less expensive than TCP. 
Since ICP produces the same region as TCP, which is what we need in light of the goal of our paper, we do not present ICP in detail. We refer the interested reader to \citep{papadopoulos2008inductive,angelopoulos2024theoreticalfoundationsconformalprediction}.
%

\subsection{Imprecise Probabilities}\label{ips}
Let $(\mathbb{Y},\Sigma_\mathbb{Y})$ be a measurable (prediction) space, where $\mathbb{Y}$ is a nonempty set and $\Sigma_\mathbb{Y}$ is a $\sigma$-algebra on $\mathbb{Y}$. A set function $\nu:\Sigma_\mathbb{Y} \rightarrow [0,1]$ is called  \textit{(Choquet) capacity} if $\nu(\emptyset)=0$, $\nu(\mathbb{Y})=1$, and $\nu(A) \leq \nu(B)$ for all $A,B\in\Sigma_\mathbb{Y}$ such that $A\subseteq B$ \citep{cella2022validity,cerreia2016ergodic,choquet1954theory}. When $\mathbb{Y}$ is not finite, $\nu$ needs to be continuous from above and below \citep[Chapter 6]{augustin2014introduction}.

Capacity $\nu$ is called a \textit{lower probability} (LP) if it is superadditive, that is, if $\nu(A \cup B) \geq \nu(A) + \nu(B)$, for all disjoint sets $A,B\in\Sigma_\mathbb{Y}$. In the remainder of the paper, we denote a generic LP by $\underline{P}$ in view of the following property. A lower probability $\underline{P}$ can be obtained as the \textit{lower envelope} of a   set $\mathcal{P}$ of probability measures on  $(\mathbb{Y},\Sigma_\mathbb{Y})$, that is,
$$
\underline{P}(A)=\inf_{P\in\mathcal{P}}P(A),
$$
for all $A\in \Sigma_\mathbb{Y}$ \citep{cella2022validity,cerreia2016ergodic,walley1991statistical}. If $\mathcal{P}$ is convex and closed,\footnote{Here and in the rest of the paper, closed has to be understood with respect to the topology endowed to the space $\Delta_\mathbb Y$ of all probability measures on $(\mathbb{Y},\Sigma_\mathbb{Y})$, typically the weak or weak$^\star$ topologies.} we call it \textit{credal set}; a credal set having a finite number of extreme elements is called a \textit{finitely generated credal set} (FGCS).\footnote{The extreme elements of an FGCS are those that cannot be written as convex combinations of one another.}

Let $\overline{P}$ denote the dual of an LP $\underline{P}$, that is, $\overline{P}(A)=1-\underline{P}(A^c)$, for all $A\in\Sigma_\mathbb{Y}$. We call $\overline{P}$ an \textit{upper probability} (UP). It is subadditive, that is, $\overline{P}(A \cup B) \leq \overline{P}(A) + \overline{P}(B)$, for all disjoint sets $A,B\in\Sigma_\mathbb{Y}$. In addition, it can be obtained as the \textit{upper envelope} of a set $\mathcal{P}$ of   probability measures on  $(\mathbb{Y},\Sigma_\mathbb{Y})$, that is, $\overline{P}(A)=\sup_{P\in\mathcal{P}}P(A)$, for all $A\in \Sigma_\mathbb{Y}$.

The duality of $\underline{P}$ and $\overline{P}$ means that knowing one of them is sufficient to retrieve the other. We may refer to either of them individually with the understanding of their one-to-one relationship \citep{gong2021judicious}.

Capacity $\nu$ is said to be \textit{$k$-monotone} if for every collection $\{A, A_1, \ldots , A_k\} \subseteq \Sigma_\mathbb{Y}$ such that $A_i \subseteq A$, for all $i \in \{1, \ldots, k\}$, we have
\begin{equation}\label{k-mon}
    \nu(A) \geq \sum_{\emptyset\neq \mathcal{I} \subseteq \{1, \ldots, k\}} (-1)^{|\mathcal{I}| -1} \, \nu \left(\bigcap_{i\in \mathcal{I}} A_i \right).
\end{equation}
If a capacity is $(k + 1)$-monotone, it is $k$-monotone as well. The smaller the $k$, the broader the class \citep{gong2021judicious}. Capacity $\nu$ is said to be \textit{$k$-alternating} if for every collection $\{A, A_1, \ldots , A_k\} \subseteq \Sigma_\mathbb{Y}$ such that $A_i \subseteq A$, for all $i \in \{1, \ldots, k\}$,
\begin{equation}\label{k-alt}
    \nu(A) \leq \sum_{\emptyset\neq \mathcal{I} \subseteq \{1, \ldots, k\}} (-1)^{|\mathcal{I}| -1} \nu(\cup_{i\in \mathcal{I}} A_i).
\end{equation}
If an LP $\underline{P}$ is a $k$-monotone capacity, then its dual UP $\overline{P}$ will be $k$-alternating. Similarly, if an UP $\overline{P}$ is a $k$-alternating capacity, then its dual LP $\underline{P}$ will be $k$-monotone.\footnote{Notice that an LP cannot be $k$-alternating because it is superadditive, and similarly a UP cannot be $k$-monotone because it is subadditive.}

Lower probability $\underline{P}$ is said to be \textit{convex} if it is a $2$-monotone capacity, that is, if $\underline{P}(A\cup B) \geq \underline{P}(A)+\underline{P}(B)-\underline{P}(A\cap B)$, for all $A,B\in\Sigma_\mathbb{Y}$. Similarly, UP $\overline{P}$ is said to be \textit{concave} if it is a $2$-alternating capacity, that is, if $\overline{P}(A\cup B) \leq \overline{P}(A)+\overline{P}(B)-\overline{P}(A\cap B)$, for all $A,B\in\Sigma_\mathbb{Y}$ \citep{caprio2023constriction}. 

Lower probability $\underline{P}$ is called a \textit{belief function} if it is an $\infty$-monotone capacity, i.e., if \eqref{k-mon} holds for every $k$; we denote it by $bel$. Upper probability $\overline{P}$ is called a \textit{plausibility function} if it is an $\infty$-alternating capacity, i.e., if \eqref{k-alt} holds for every $k$; we denote it by $pl$. Belief and plausibility functions are dual \citep{shafer1976mathematical}. A belief function induces a set-valued random variable on the power set of $\mathbb{Y}$ \citep{gong2021judicious}: if $bel$ is a belief function, its associated \textit{mass function} is the non-negative set function
\begin{equation}\label{mass_funct}
    m:\Sigma_\mathbb{Y} \rightarrow [0,1] , \quad A \mapsto m(A)\coloneqq\sum_{B\subseteq A} (-1)^{|A-B|} bel(B),
\end{equation}
where $A-B \equiv A \cap B^c$, and the subsets $B$ of $A$ have to belong to $\Sigma_\mathbb{Y}$ as well. 
Mass function $m$ has the following properties: 
\begin{itemize}
  \setlength\itemsep{0.5em}
  \setlength{\itemindent}{0.5cm}
    \item[(a)] $m(\emptyset)=0$;
    \item[(b)] $\sum_{B\subseteq\mathbb{Y}} m(B)=1$;
    \item[(c)] $bel(A)=\sum_{B\subseteq A} m(B)$, and this representation is unique to $bel$.
\end{itemize}
Equation \eqref{mass_funct} is called the Möbius transform of $bel$ \citep{yager2008classic}. A mass function $m$ induces a   probability measure on $\Sigma_\mathbb{Y}$, namely as the distribution of a random set. These concepts are further studied by \citet{gong2021judicious}.

Plausibility function $pl$ is said to be \textit{consonant} if there exists a function $\pi:\mathbb{Y}\rightarrow [0,1]$ such that 
\begin{itemize}
  \setlength\itemsep{0.5em}
  \setlength{\itemindent}{0.5cm}
    \item[(i)] $\sup_{y\in\mathbb{Y}}\pi(y)=1$;
    \item[(ii)] $pl(A)=\sup_{y\in A} \pi(y)$, $A\in \Sigma_\mathbb{Y}$.
\end{itemize}
Function $\pi$ corresponds to the probability density or mass function (pdf/pmf) of $m$, and is called a \textit{plausibility contour}. We now show an interesting algebraic property of consonant plausibility functions.

Recall that a {\em semigroup} is a generic set $S$ together with a binary operation $\dagger$ (that is, a function $\dagger: S \times S \rightarrow S$) that satisfies the associative property, i.e. $(a \dagger b) \dagger c = a \dagger (b \dagger c)$, for all $a,b,c \in S$. Semigroup $S$ is called a {\em monoid} if it has an identity element $e$, that is, an element such that $a \dagger e = e \dagger a =a$, for all $a \in S$. Recall also that a {\em monoid homomorphism} is a map between monoids that preserves the monoid operation and maps the identity element of the first monoid to that of the second monoid. We have the following.

\begin{lemma}\label{lem-monoid}
A consonant plausibility function $pl$ is a monoid homomorphism between the monoids $(\Sigma_{\mathbb{Y}}, \cup)$ and $([0, 1], \oplus)$, where $\cup$ is the set union operation and $\oplus$ is the tropical addition on $[0, 1]$
\end{lemma}

\begin{proof}
    Consider the semigroups $(\Sigma_\mathbb{Y}, \cup)$ and $([0,1],\oplus)$, where $\cup$ is the usual set union operation, and $\oplus$ is the tropical addition, i.e. $a \oplus b = \max\{a,b\}$, for all $a,b \in [0,1] \subset \mathbb{R}$. They are both monoids with identity elements $\emptyset$ and $0$, respectively. 
    
    Then, a consonant plausibility function $pl$ is a map $pl: \Sigma_\mathbb{Y} \rightarrow [0,1]$ such that $pl(\emptyset)=0$, and $pl(A \cup B)=pl(A) \oplus pl(B)$. This shows that $pl$ is a monoid homomorphism, as desired.
\end{proof}

While at the moment Lemma \ref{lem-monoid} seems like a result of independent interest, unrelated to our main result in Section \ref{narrower}, we point out how in the future it may be useful to relate Imprecise Probabilities to an algebraic version of Conformal Prediction, in the spirit of the field of Algebraic Statistics \citep{sullivant}.

Note that $\pi(y)$ is the upper probability assigned to the singleton set $\{y\}$, which
means that the entire lower and upper probability pair $(\underline{P},\overline{P})$ is determined by the plausibility assigned to singletons. This is a unique feature of the consonance model \citep{cella2022validity,dubois1990consonant}. If we denote by $\mathcal{F}_m$ the set of all focal elements, i.e., $A \in \Sigma_\mathbb{Y}$ such that $m(A) > 0$, then consonance corresponds to the case of having nested focal elements. Specifically, for any two elements $A, B \in \Sigma_\mathbb{Y}$ we have either $A \subseteq B$ or $B \subseteq A$, indicating a hierarchical structure \citep{shafer1976mathematical}. We mention in passing that in classical possibility theory, ``consonance'' refers to the nestedness property we just described, while condition $\sup_{y\in\mathbb{Y}}\pi(y)=1$ is usually referred to as ``normalization''. We do not make a distinction between the two to be consistent with the terminology in Cella and Martin's works, and because normalization combined with the condition $pl(A)=\sup_{y\in A} \pi(y)$, $A\in \Sigma_\mathbb{Y}$, implies nestedness.

We conclude by pointing out that Imprecise Probabilistic concepts like the ones we presented in this section are routinely implemented in many applied fields, such as Machine Learning and Artificial Intelligence \citep{denoeux,zaffalon,dest1,inn,ibcl,credal-learning,second-order,pmlr-v216-wimmer23a,pmlr-v216-sale23a}, Statistics \citep{chau2024credaltwosampletestsepistemic,RODEMANN2024112186,ext_prob,novel_bayes,dipk,ergo.th,impr-MSG,caprio2025optimaltransportepsiloncontaminatedcredal}, Engineering \citep[Chapter 4]{engin} -- including aerospace engineering \citep[Chapter 5]{engin2} -- and Economics \citep{boppi,denk2,DENK}.

\section{Relationship between Belief Functions and TCP}\label{rel1}
This section is based on \citet[Section 4.2]{cella2022validity}, who highlight the relationship between conformal transducers and plausibility contours. Further works that inspect such a relationship are \citep{alireza2,alireza,eyke2,bordini,caprio-conformalized}.

One of the main concerns in the conformal literature is to identify function $\pi$, taking values in $[0,1]$, such that $\pi(Y_{n+1},\mathbf{Y^n})$ is stochastically no smaller than $\text{Unif}(0,1)$ under any exchangeable $P\in\Delta_\mathbb{Y}$. 
Also compare this to classical hypothesis testing in statistics, where the p-values under a true null hypothesis are distributed uniformly. 
\citet[Theorem 11.1]{vovk2005algorithmic} show that the conformal transducer returned by Algorithm \ref{algo1} satisfies this requirement. Suppose now that function $\pi$ also satisfies 
\begin{equation}\label{eq_imp_1}
    \sup_{\tilde{y}\in\mathbb{Y}}\pi(\tilde{y},\mathbf{y^n})=1 \, , \quad \text{ for all } \mathbf{y^n} \in \mathbb{Y}^n.
\end{equation}
This property holds quite generally for conformal prediction in continuous-data problems: if $\tilde y$ is a point at which the minimum of $\tilde{y}\mapsto \Psi(\mathbf{y^n},\tilde{y})$ is achieved, then $\pi(\tilde y,\mathbf{y^n})=1$ and \eqref{eq_imp_1} holds. For discrete-data problems, see \citet{cella2021valid}. Notice that consonance can also be satisfied with a transformation of the conformal transducer $\pi$. For example, in \citet[Section 7]{cella2021valid}, the authors suggest the following two adjusted conformal transducers: $\pi^\prime (y,\mathbf{y^n}) = \pi(y,\mathbf{y^n})/\sup_y \pi(y,\mathbf{y^n})$, for all $y\in\mathbb{Y}$, and $$\pi^{\prime\prime} (y,\mathbf{y^n}) = \begin{cases}
    1 & \text{if } y \in \text{argsup}_{y\in\mathbb{Y}} \pi(y,\mathbf{y^n})\\
    \pi (y,\mathbf{y^n}) & \text{otherwise}
\end{cases},$$
with the latter being more efficient. From \eqref{eq_imp_1}, we can define a (predictive) upper probability as follows\footnote{Let us note in passing that, when defining upper probability on the basis of the plausibility contour, we turn ordinal (order) information into cardinal information.}

\begin{equation}\label{eq_imp_2}
    \overline{\Pi}_{\mathbf{y^n}}(A)=\sup_{\tilde{y}\in A} \pi(\tilde{y},\mathbf{y^n}), \quad A\in \Sigma_\mathbb{Y}.
\end{equation}

It is immediate to see how the upper probability in \eqref{eq_imp_2} is a consonant plausibility function, as introduced in Section \ref{ips}. Conformal transducer $\pi$ corresponds to a plausibility contour, and it fully determines the upper and lower probabilities of the elements of $\Sigma_\mathbb{Y}$ via \eqref{eq_imp_2} and the duality property of lower probabilities. 

We now discuss some interesting properties of upper probability $\overline{\Pi}_{\mathbf{y^n}}$, namely that it is coherent, supremum preserving, and tropically finitely additive.

Recall that the tropical semiring is the $3$-tuple $(\mathbb{R}\cup\{\infty\},\oplus,\otimes)$, where, for all $a,b\in\mathbb{R}\cup\{\infty\}$, $a \oplus b = \max\{a,b\}$, and $a \otimes b = a + b$ \citep{maclagan}. As a consequence, we can write that, for all $A,B \in \Sigma_\mathbb{Y}$, $\overline{\Pi}_{\mathbf{y^n}}(A) \oplus \overline{\Pi}_{\mathbf{y^n}}(B) = \max\{\overline{\Pi}_{\mathbf{y^n}}(A),\overline{\Pi}_{\mathbf{y^n}}(B)\}$ and $\overline{\Pi}_{\mathbf{y^n}}(A) \otimes \overline{\Pi}_{\mathbf{y^n}}(B)= \overline{\Pi}_{\mathbf{y^n}}(A) + \overline{\Pi}_{\mathbf{y^n}}(B)$.

Recall also that a generic upper probability $\overline{P}$ is supremum preserving if $\overline{P}(\cup_{A\in\mathcal{A}} A) = \sup_{A\in\mathcal{A}} \overline{P}(A)$, for any possible collection $\mathcal{A}\subseteq \Sigma_\mathbb{Y}$ \citep{DECOOMAN1999173}. 

Finally, we say that a generic upper probability $\overline{P}$ is \textit{coherent} à la \citet[Section 2.5]{walley1991statistical} if its dual lower probability $\underline{P}$ is such that, given an arbitrary collection $\mathscr{K}$ of bounded random variables on $\mathbb{Y}$, we have that $\sup[\sum_{j=1}^n (X_j-\underline{P}(X_j)) -m (X_0-\underline{P}(X_0))] \geq 0$, whenever $m,n\in\mathbb{Z}_+$, and $X_0,X_1,\ldots,X_n$ (not necessarily distinct) are in $\mathscr{K}$. 

\begin{lemma}[Properties of a Consonant Upper Probability]\label{prop_tropical}
    Let $\mathcal{A} \subseteq \Sigma_\mathbb{Y}$ be a generic collection of subsets of $\mathbb{Y}$. Upper probability $\overline{\Pi}_{\mathbf{y^n}}$ defined in \eqref{eq_imp_2} is supremum preserving and coherent à la Walley. In addition,  $\overline{\Pi}_{\mathbf{y^n}}$ is tropically finitely additive, that is,
    \begin{equation*}
    |\mathcal{A}|<\infty \implies \overline{\Pi}_{\mathbf{y^n}}\left(\bigcup_{A\in\mathcal{A}} A\right) = \bigoplus_{A\in\mathcal{A}} \overline{\Pi}_{\mathbf{y^n}}(A).
    \end{equation*}
\end{lemma}

\begin{proof}
    Upper probability $\overline{\Pi}_{\mathbf{y^n}}$ is supremum preserving and coherent à la Walley by \citet{DECOOMAN1999173} and \citet[Section 4.6.1]{augustin2014introduction}. Tropical finite additivity follows from  $\overline{\Pi}_{\mathbf{y^n}}$ being supremum preserving, having assumed $|\mathcal{A}|<\infty$, and the definition of additivity in the tropical semiring. It can also be derived immediately from Lemma \ref{lem-monoid}, or the ``maxitivity'' of possibility theory \citep{Dubois2000-book}.
\end{proof}

Unlike classical finite additivity, tropical finite additivity does not require the events in $\mathcal{A}$ to be disjoint. The practical implications of Lemmas \ref{lem-monoid} and \ref{prop_tropical} -- especially the algebraic properties and the tropical additivity of consonant plausibility functions -- will be the subject of future work.

\begin{remark}\label{remark_notation}
    Let us pause here to discuss a slight notational abuse that we perpetrate throughout the paper. When we say that ``conformal transducer $\pi$ corresponds to a plausibility contour'', we mean that conformal transducer $\pi(\cdot,\mathbf{y^n})$ corresponds to a plausibility contour $\pi(\cdot)$, for all $\mathbf{y^n}\in\mathbb{Y}^n$. 
\end{remark}

From a practical point of view, adding consonance to conformal prediction (that is, requiring \eqref{eq_imp_1} and \eqref{eq_imp_2} to hold) creates no new computational challenges. The standard use of Algorithm \ref{algo1}’s output is to extract the prediction region $\mathscr{R}_\alpha(\mathbf{y^n})$ which is just the collection of all $ \tilde{y}$'s such that $\pi(\tilde{y},\mathbf{y^n})$ exceeds $\alpha$. 

With the addition of consonance, \citet{cella2022validity} recommend two additional summaries. First, at least in low-dimensional problems, a plot of $\tilde{y}\mapsto\pi(\tilde{y},\mathbf{y^n})$ to give a visual assessment of the information available in the data $\mathbf{y^n}$ regarding $Y_{n+1}$, similar to the Bayesian posterior predictive density function \citep[Section 6]{cella2022validity}. Second, for any $A\in \Sigma_\mathbb{Y}$, the (prediction) upper probability at $A$ can be approximated as
$$\overline{\Pi}_{\mathbf{y^n}}(A) \approx \max_{\tilde{y}\text{ on the grid and in } A}\pi(\tilde{y},\mathbf{y^n}).$$
Consonance also induces several practically relevant properties that are inspected in \citet[Section 4.2]{cella2022validity} and in \citet[Chapter 4]{augustin2014introduction}. 

\section{Imprecise Highest Density Regions}\label{ihdr_sec}

This section introduces the concept of Imprecise Highest Density Region (IHDR), a subset of $\mathbb Y$ obtained from a methodology that employs techniques from the imprecise probability literature \citep{coolen1992imprecise}. Consider again a process $Y_1,Y_2,\ldots$ of random elements taking values in $\mathbb{Y}$. 

Like in the case of conformal prediction, the goal is to predict $Y_{n+1}$ using a method that is valid. To this end, an IHDR proceeds from a credal set $\mathcal{P}_\text{pred}$ of candidate distributions on $\mathbb{Y}$, which is supposed to be reliable in the sense that it contains the true distribution.\footnote{Subscript ``pred'' denotes the fact that, in the present paper, we are interested in a credal set of predictive distributions derived from the data $\mathbf{y^n}$ at hand via some procedure, e.g. generalized Bayes' or geometric updating \citep{gong2021judicious}.} 

\begin{definition}[Imprecise Highest Density Region \citep{coolen1992imprecise}]\label{ihdr-def}
	Let $\mathcal{P}_\text{pred}$ be a credal set of distributions on $\mathbb{Y}$, and $\underline{P}$ be its lower probability. Call $Y_{n+1}$ a random quantity that takes values in $\mathbb{Y}$, and $\alpha$ any value in $[0,1]$. Then, set $\text{IR}_\alpha \subseteq \mathbb{Y}$ is called a $(1-\alpha)$-\textit{Imprecise Highest Density Region} (IHDR) if 
 $$
 \underline{P}  \big[  Y_{n+1} \in \text{IR}_\alpha  \big]= 1-\alpha
 $$ 
 and $\int_{\text{IR}_\alpha} \text{d}y$ is a minimum. If $\mathbb Y$ is at most countable, we replace $\int_{\text{IR}_\alpha} \text{d}y$ with $|\text{IR}_\alpha|$.
\end{definition}

As a consequence of Definition \ref{ihdr-def}, we have that $\text{IR}_\alpha$ is the smallest subset of $\mathbb Y$ such that 

\begin{equation}\label{guar-ihdr}
    {P}[Y_{n+1} \in \text{IR}_\alpha]\geq 1-\alpha, \quad \forall P \in \mathcal{P}_\text{pred}.
\end{equation}
Here lies the appeal of the IHDR concept. Contrary to the guarantee in \eqref{eq_imp6} though -- which holds for all the possible exchangeable distributions $P$ on $\mathbb{Y}$ -- the one in \eqref{guar-ihdr} holds only for the distributions that are plausible for the analysis at hand, that is, for all those in $\mathcal{P}_\text{pred}$.

{\bf Example.} Let us now give an illustrating example of how to derive an IHDR. Credal set $\mathcal{P}_\text{pred}$ may originate from different sources. One instance is 
robust Bayesian inference \citep{berger2,hartigan,caprio2024credal}, where $\mathcal{P}_\text{pred}$ corresponds to the (posterior) predictive credal set induced by a prior credal set. 
Suppose that we observe the first $n$ elements of the process $\{Y_k\}_{k \in \mathbb{N}}$. A traditional Bayesian learner would first specify a likelihood for the observations, and then elicit a prior on the parameter of the likelihood distribution. 
Imagine that the $Y_k$'s are distributed according to a Poisson distribution $\text{Pois}(\lambda)$, so that the conjugate prior for the rate parameter $\lambda$ is a Gamma distribution. Suppose then that prior information is available but comes from different sources or experiments. For instance, assuming $\lambda \sim \text{Gamma}(a_j, b_j)$ may align with experiments on one group $j \in \{1, \ldots, J \}$. 
Call $P_j$ the probability measure whose pdf is $\text{Gamma}(a_j, b_j)$, $j \in \{1, \ldots, J \}$. 
In this case, a prior finitely generated credal set (FGCS)
\begin{align*}
\mathcal{P}_\text{prior} = \left\lbrace{Q: Q = \sum_{j=1}^J \beta_j \cdot P_j \text{, } \beta_j \geq 0 \text{, } \sum_{j=1}^J \beta_j = 1 }\right\rbrace
\end{align*}
is a natural choice. This corresponds to the Bayesian sensitivity analysis (BSA) approach to inference  \citep{berger1984robust,ellsberg1961risk,gilboa2016ambiguity}, \citep[Section 5.9]{walley1991statistical}. The extreme elements of $\mathcal{P}_\text{prior}$ are $\text{ex}\mathcal{P}_\text{prior} = \{P_1, \ldots , P_J\}$. By computing their posterior, we obtain the extreme elements of the posterior FGCS, $\text{ex}\mathcal{P}_\text{post} \subseteq \{P_1^\text{post}, \ldots , P_J^\text{post}\}$ (continuing our example, the elements of $\text{ex}\mathcal{P}_\text{post}$ are Gamma distributions with updated shape and rate parameters). 

The goal now is to predict $Y_{n+1}$ using a valid or reliable method. 
%
To do so, first the agent derives the extrema of the predictive FGCS, $\text{ex}\mathcal{P}_\text{pred}\subseteq\{P_1^\text{pred},\ldots,P_J^\text{pred}\}$. These are the probability measures whose pdf's are computed as $p_j^\text{pred}(y_{n+1} | y_1, \ldots, y_n) = \int_0^\infty \ell(y_{n+1} | \lambda) \cdot p_j^\text{post} (\lambda | y_1, \ldots, y_n) \text{d}\lambda$, $ j \in \{1, \ldots, J\}$, where $\ell(y_{n+1} | \lambda)$ is our Poisson likelihood, and $p_j^\text{post}$ is the pdf of the posterior measure $P_j^\text{post}$, $j \in \{1, \ldots, J\}$. By well-known results in Bayesian inference \citep{hoff}, each such a predictive distribution is a Negative Binomial. The predictive FGCS is given by $\mathcal{P}_\text{pred} = \text{Conv}(\text{ex}\mathcal{P}_\text{pred}) \subseteq \Delta_{\mathbb{Y}}$, where $\text{Conv}(\cdot)$ denotes the convex hull operator. By \citet[Proposition 3]{caprio2024credal}, we know that $\underline{P}(A)=\inf_{P \in \text{ex}\mathcal{P}_\text{pred}}P(A)=\inf_{P \in \mathcal{P}_\text{pred}}P(A)$, for all $A\in\Sigma_\mathbb{Y}$. Hence, to find our desired IHDR, we only have to focus on searching for the smallest region $\text{IR}_\alpha$ such that \eqref{guar-ihdr} is satisfied for all $P \in \text{ex}\mathcal{P}_\text{pred}$. Notice that the complexity of computation is linear in $|\text{ex}\mathcal{P}_\text{pred}|$, that is, in the number of extreme elements of $\mathcal{P}_\text{pred}$ \citep[Section 9.2.1]{augustin2014introduction}.
\hfill $\blacktriangle$

Oftentimes, in Imprecise Probability theory, it is posited that
\begin{equation}\label{assumption_1}
    \mathbf{P} \in \mathcal{P}_\text{pred},
\end{equation}
where -- with an abuse of notation -- $\mathbf{P}$ has to be understood as the marginal probability measure on $Y_{n+1}$ derived from the true joint distribution of $Y^\infty$.\footnote{Throughout the paper, we denote by $\mathbf{P}$ both the true data generating process on $Y^\infty$ and its marginal on $Y_{n+1}$. This abuse of notation allows us to better communicate our results, and does not induce confusion, as it is obvious from the context whether we are referring to the joint or the marginal distributions.} This assumption is not overly strong, as it may appear at first glance. The credal set $\mathcal{P}_\text{pred}$ will generally be wider, the higher the uncertainty around the true $\mathbf{P}$. There is also an active literature that studies statistical tests which check whether credal sets are calibrated à la \eqref{assumption_1} \citep{acharya2015optimal,gao2018robust,mortier2023calibration,liu2024robustness,chau2024credaltwosampletestsepistemic}.

Assumption \eqref{assumption_1} is not needed for $\mathcal{M}(\overline{\Pi}_\mathbf{y^n})$. Indeed,   \citet{martin2022valid} shows that $\overline{\Pi}_\mathbf{y^n}$ is the minimal outer consonant approximation of the true data generating process $\mathbf{P}$. This means that $\overline{\Pi}_\mathbf{y^n}$ is the narrowest upper bound for the true distribution, that also satisfies the consonance assumption. We can then conclude -- although it is not explicitly shown in their work -- that credal region $\mathcal{M}(\overline{\Pi})$ contains the true data generating process.

{\bf Possible Criticism to a Generic IHDR.} 
It requires the agent to come up with $\mathcal{P}_\text{pred}$. To counter this objection, we point out how modeling choices are crucial to model-based inference. A priori, we cannot say that model-based approaches are to be preferred to model-free ones, or vice versa. The choice will depend on the available information. Let us also remark that conformal prediction is not a completely model-free approach either, since the user is required to specify a non-conformity measure $\Psi$, which has a strong influence on the size of the conformal prediction region.

\section{Building the Equivalent Prediction Region}\label{narrower}
In this section, we present the main contribution of our work. We show that -- for a given nonconformity measure $\Psi$ used to derive the conformal transducer $\pi$ -- credal set $\mathcal{M}(\overline{\Pi}_\mathbf{y^n})$ can be used to obtain a prediction region that is equivalent to the CPR, and that retains the same uniform guarantees. 

Suppose (i) that the process $Y_1,Y_2,\ldots$ is exchangeable and governed by a unique distribution $\mathbf{P}$ (so that distribution shifts are not allowed), and (ii) that the conformal transducer $\pi$ is consonant, that is, $\sup_{\tilde{y}\in\mathbb{Y}}\pi(\tilde{y},\mathbf{y^n})=1$. \\
\indent Let $\overline{\Pi}_\mathbf{y^n}(A)=\sup_{\tilde{y}\in A} \pi(\tilde{y},\mathbf{y^n})$, for all $A \in\Sigma_\mathbb{Y}$, be the upper probability induced by $\pi$. Let also $\underline{\Pi}_\mathbf{y^n}(A)=1-\overline{\Pi}_\mathbf{y^n}(A^c)$, for all $A \in\Sigma_\mathbb{Y}$, be the lower probability dual to $\overline{\Pi}_\mathbf{y^n}$. Let $\mathcal{M}(\overline{\Pi}_\mathbf{y^n})$ be the credal set induced by $\overline{\Pi}_\mathbf{y^n}$, that is,
$$\mathcal{M}(\overline{\Pi}_\mathbf{y^n})\coloneqq \left\lbrace{P: P(A) \leq \overline{\Pi}_\mathbf{y^n}(A) \text{, } \forall A \in \Sigma_\mathbb{Y}}\right\rbrace.$$
To see that $\mathcal{M}(\overline{\Pi}_\mathbf{y^n})$ is indeed a credal set, we refer the reader to \citet{marinacci2004introduction,walley1991statistical}.
We denote by $\text{IR}_\alpha^\mathcal{M}$, $\alpha\in [0,1]$, the IHDR associated with the credal set $\mathcal{M}(\overline{\Pi}_\mathbf{y^n})$ induced by the conformal procedure, that is, $\text{IR}_\alpha^\mathcal{M} \subseteq \mathbb{Y}$ such that $\underline{\Pi}_\mathbf{y^n}(\text{IR}_\alpha^\mathcal{M})= 1-\alpha$ and $\int_{\text{IR}_\alpha^\mathcal{M}}\text{d}y$ is minimal. Then, we have the following.

\begin{proposition}[IHDR Corresponds to the Conformal Prediction Region]\label{nar-pr}
    For any $\alpha\in [0,1]$ and any $n\in\mathbb{N}$, the following is true
    $$\text{IR}_\alpha^\mathcal{M} = \mathscr{R}_\alpha(\mathbf{y^n}).$$
\end{proposition}

\begin{proof}

    Consider the function $\gamma: \mathbb{Y} \rightarrow [0,1]$, 
    $$y \mapsto \gamma(y)=\begin{cases}
        \pi(y) &\text{if } \pi(y) \leq 0.5\\
        1-\pi(y) &\text{if } \pi(y) > 0.5
    \end{cases}.$$
    It is easy to see that $\gamma(y) \leq \pi(y)$, for all $y \in \mathbb{Y}$. In addition, by the consonance property of $\pi$, there exists $\tilde y \in \mathbb Y$ such that $\gamma(\tilde y)=0$. In turn, we have that $[\gamma,\pi]$ is a cloud \citep[Definition 4.6]{augustin2014introduction}. 
    Neumaier's probabilistic constraint on clouds \citep{neumaier}, \citep[Equation (4.9)]{augustin2014introduction} gives us the following
    \begin{align*}
        P[Y_{n+1}&\in\mathscr{R}_\alpha(\mathbf{y^n})]\\
        &= P[Y_{n+1}\in \{y\in\mathbb{Y} : \pi(y,\mathbf{y^n}) > \alpha\}]\\
        &\geq 1-\alpha = \underline{\Pi}_\mathbf{y^n}(Y_{n+1}\in \text{IR}_\alpha^\mathcal{M}),
    \end{align*}
    for all $P\in\mathcal{M}(\overline{\Pi}_\mathbf{y^n})$.
    In turn, we have
    \begin{align}\label{ineq-incl}
    \begin{split}
        \underline{\Pi}_\mathbf{y^n}[Y_{n+1}&\in\mathscr{R}_\alpha(\mathbf{y^n})]\\
        &= \underline{\Pi}_\mathbf{y^n}[Y_{n+1}\in \{y\in\mathbb{Y} : \pi(y,\mathbf{y^n}) > \alpha\}]\\
        &\geq 1-\alpha = \underline{\Pi}_\mathbf{y^n}(Y_{n+1}\in \text{IR}_\alpha^\mathcal{M}).
    \end{split}
    \end{align}
    By the monotonicity of lower probabilities (see Section \ref{ips}), then, we can conclude that $\text{IR}_\alpha^\mathcal{M} \subseteq \mathscr{R}_\alpha(\mathbf{y^n})$. But, by the definition of Conformal Prediction Region, $\text{IR}_\alpha^\mathcal{M}$ cannot be strictly included in $\mathscr{R}_\alpha(\mathbf{y^n})$. In turn, this implies that $\text{IR}_\alpha^\mathcal{M} = \mathscr{R}_\alpha(\mathbf{y^n})$, as desired.
\end{proof}

Proposition \ref{nar-pr} ensures us of the following. Once we derive the conformal transducer $\pi$, if we take the imprecise probabilistic route based on (predictive) credal set $\mathcal{M}(\overline{\Pi}_\mathbf{y^n})$, we are able to obtain a prediction region (an IHDR) $\text{IR}_\alpha^\mathcal{M}$. The latter corresponds to the Conformal Prediction Region $\mathscr{R}_\alpha(\mathbf{y^n})$, for all possible choices of significance level $\alpha\in [0,1]$. 
We also point out how the proof of Proposition \ref{nar-pr} links Conformal Prediction to the IP concept of a cloud, another profound connection between the two literatures.

We note in passing that another proof of Proposition \ref{nar-pr} exists, and is based on \citet[Page 25]{martin2022valid} as follows. Thanks to the consonance of $\pi$, we have that
\begin{align}\label{str-alpha-cut}
\begin{split}
    \mathscr{R}_\alpha(\mathbf{y^n})&=\{y\in\mathbb{Y} : \pi(y,\mathbf{y^n}) > \alpha\} \\ &= \bigcap \left\lbrace{A \in \Sigma_\mathbb{Y} : \underline{\Pi}_\mathbf{y^n}(A) \geq 1-\alpha}\right\rbrace \\ &= \text{IR}_\alpha^\mathcal{M},
\end{split}
\end{align}
for all $\alpha\in [0,1]$. 

A natural question at this point is whether $\text{IR}_\alpha^\mathcal{M}$ enjoys the same probabilistic guarantees as $\mathscr{R}_\alpha(\mathbf{y^n})$. The following proposition and corollary provide us with a positive answer.

\begin{proposition}[Elements of the IHDR]\label{prop-guar-imp}
    For all $\alpha\in (0,1]$, $P\in\mathcal{M}(\overline{\Pi}_\mathbf{y^n})$ if and only if $P[Y_{n+1}\in\{y\in\mathbb{Y} : \pi(y,\mathbf{y^n}) >\alpha\}]\geq 1-\alpha$.\footnote{The set $\{y\in\mathbb{Y} : \pi(y,\mathbf{y^n}) >\alpha\}$ is called a \textit{strong $\alpha$-cut} of $\pi$. If we replace the strong inequality with a weak one, the set is called a \textit{regular $\alpha$-cut} of $\pi$ \citep[Section 4.6]{augustin2014introduction}.}
\end{proposition}

\begin{proof}
    Immediate from \citet[Proposition 4.1]{augustin2014introduction}.
\end{proof}

\begin{corollary}[Probabilistic Guarantees of the IHDR]\label{cor-guar-imp}
    For all $\alpha\in [0,1]$, the following holds uniformly in $P$ and in $n$,\footnote{Here too ``uniformly in $P$'' has to be understood as ``for all exchangeable $P$ on $\mathbb{Y}$''.}
    $$P[Y_{n+1}\in \text{IR}_\alpha^\mathcal{M}]\geq 1-\alpha.$$
\end{corollary}

\begin{proof}
    Pick any $\alpha\in (0,1]$ and any $n\in\mathbb{N}$. Then, by \citet[Theorem 11.1]{vovk2005algorithmic} and \citet[Section 4.2]{cella2022validity}, we have that
    \begin{equation}\label{eq-v-imp}
        P[Y_{n+1}\in\{y\in\mathbb{Y} : \pi(y,\mathbf{y^n}) >\alpha\}]\geq 1-\alpha,
    \end{equation}
    uniformly in $P$, that is, for all exchangeable distribution $P$ on $\mathbb{Y}$. By Proposition \ref{prop-guar-imp}, then, all the (exchangeable) probabilities $P\in \Delta_\mathbb{Y}$ satisfying \eqref{eq-v-imp} belong to the credal set $\mathcal{M}(\overline{\Pi}_\mathbf{y^n})$. We also have that, by definition, $\underline{\Pi}_\mathbf{y^n}[Y_{n+1}\in \text{IR}_\alpha^\mathcal{M}]= 1-\alpha$, which implies that $P[Y_{n+1}\in \text{IR}_\alpha^\mathcal{M}]\geq 1-\alpha$, for all $P\in \mathcal{M}(\overline{\Pi}_\mathbf{y^n})$. 
    
    For $\alpha=0$, the statement follows from \eqref{str-alpha-cut} and \citet[Theorem 11.1]{vovk2005algorithmic}. Notice how we could have directly used \citet[Theorem 11.1]{vovk2005algorithmic} to prove the whole result (without splitting in the $\alpha \in (0,1]$ and $\alpha=0$ cases), but we preferred to state it as a corollary to Proposition \ref{prop-guar-imp} to better study the relationship between conformal prediction and imprecise probabilities.
\end{proof}
By combining Proposition \ref{nar-pr} and Corollary \ref{cor-guar-imp} together, we can conclude that an imprecise probabilistic approach delivers the same prediction region as classical Transductive Conformal Prediction,
and it retains the same uniform probabilistic guarantees. 

Let us also point out that Proposition \ref{nar-pr} is an immediate consequence of \citet[Theorem 2]{couso_cuts}, where it is also shown that Corollary \ref{cor-guar-imp} holds for all $P \in \mathcal{M}(\overline{\Pi}_\mathbf{y^n})$. Our result generalizes it slightly, by showing that the statement actually holds for all exchangeable $P$ on $\mathbb{Y}$.

\subsection{Varying the Nonconformity Measure $\Psi$}
Proposition \ref{nar-pr} and Corollary \ref{cor-guar-imp} show that, {\em once we fix the non-conformity measure $\Psi$}, an imprecise-probabilistic approach gives us a prediction region that is equivalent to the classical CPR. In the next paragraphs, we will study what happens when we change $\Psi$.

Notice that, for any $n\in\mathbb{N}$ representing the cardinality of the (training) dataset, Conformal Prediction (CP) can be seen as a set-valued function 
\begin{align}\label{cp-function}
\begin{split}
    CP:[0,1] \times \mathbb{Y}^n \times \mathscr{F} &\rightarrow \Sigma_\mathbb{Y},\\
    (\alpha,\mathbf{y^n},\Psi) &\mapsto CP(\alpha,\mathbf{y^n},\Psi)=\mathscr{R}^\Psi_\alpha(\mathbf{y^n}).
\end{split}
\end{align}
Here $\mathscr{F}\subseteq \mathbb{R}^{\mathbb{Y}^{n+1}}$ denotes the set of all possible non-conformity measures, 
\begin{align}\label{coll-cons}
    \mathscr{F}\coloneqq \left\lbrace{\Psi: \mathbb{Y}^{n} \times \mathbb{Y} \rightarrow \mathbb{R}: (\mathbf{y}_{-i}^{\mathbf{n+1}},y_i) \mapsto \psi_i(\mathbf{y^{n+1}})\in\mathbb{R}}\right\rbrace,
\end{align}
and $\mathscr{R}^\Psi_\alpha(\mathbf{y^n}) \equiv \mathscr{R}_\alpha(\mathbf{y^n}) \in \Sigma_\mathbb{Y}$ denotes the Conformal Prediction Region (in previous results we omitted the explicit reference to the choice of non-conformity measure $\Psi$ for notational convenience).\footnote{We make the implicit assumption that $\mathscr{R}^\Psi_\alpha(\mathbf{y^n}) \in \Sigma_\mathbb{Y}$. To relax it, simply substitute $\Sigma_\mathbb{Y}$ with $2^\mathbb{Y}$.} 

\begin{proposition}[Refining a CPR]\label{cp_vs_ip_nice}
    Pick any $\alpha \in [0,1]$ and any $\Psi\in\mathscr{F}$. 
    Then, there exists $\Psi^\prime \neq \Psi$
    such that $\text{IR}^\mathcal{M}_\alpha = \mathscr{R}_\alpha^{\Psi}(\mathbf{y^n}) \supseteq \mathscr{R}_\alpha^{\Psi^\prime}(\mathbf{y^n})$, and the inclusion is strict for some value of $\alpha$. 
    In addition, $P[Y_{n+1}\in \mathscr{R}_\alpha^{\Psi^\prime}(\mathbf{y^n})]\geq 1-\alpha$, uniformly in $P$ and in $n$.\footnote{Here too ``uniformly'' has to be understood in the sense of ``for all the exchangeable probability measures on $\mathbb{Y}$''.}
\end{proposition}

\begin{proof}
    The first part is an immediate consequence of \citet[Theorem 2.10]{vovk2005algorithmic}. There, the authors provide a proof by construction, that can be used to derive $\mathscr{R}_\alpha^{\Psi^\prime}(\mathbf{y^n})$, and in turn the new non-conformity measure $\Psi^\prime$. The uniform probabilistic guarantee follows from \citet[Theorem 1.11]{vovk2005algorithmic}.
\end{proof}
Proposition \ref{cp_vs_ip_nice} tells us that, while it is true that the
CPR $\mathscr{R}_\alpha^\Psi(\mathbf{y^n})$ corresponds to the IHDR $\text{IR}^\mathcal{M}_\alpha$ associated with the conformally-built credal set $\mathcal{M}(\overline{\Pi}_\mathbf{y^n})$ for a fixed non-conformity measure $\Psi$, we can always find a new non-conformity measure $\Psi^\prime$ that further improves on $\text{IR}^\mathcal{M}_\alpha = \mathscr{R}_\alpha^{\Psi}(\mathbf{y^n})$. In principle, different nonconformity measures may reduce the size of the credal set; in addition, further shrinking the set may eventually require more data.

\subsection{Discussion}\label{disc}
Before concluding this section, let us add a discussion of the results we presented so far. As pointed out by Ryan Martin and coauthors in their recent works (see e.g. \citep{cella2023possibility,martin2023fisher,martin2022valid}, and especially \citep{cella2021valid}), there seems to be a sense in which the conformal prediction procedure assumes a vacuous prior, which captures the concept of maximal (prior) ambiguity. That is, conformal prediction appears to be akin to a  Bayesian Sensitivity Analysis (BSA) procedure \citep{berger1984robust} whose prior credal set is the whole space of parameter probabilities. In BSA, though, the posterior parameter probabilities (and hence, also the predictive probabilities on $\mathbb{Y}$) obtained from this prior class are again vacuous: no finite sample is enough to annihilate a sufficiently extreme prior belief \citep{pericchi1998sets}, \citep[Appendix B]{caprio2024credal}. On the contrary, conformal prediction is able to derive a non-vacuous predictive credal set $\mathcal{M}(\overline{\Pi}_\mathbf{y^n}) \subsetneq \Delta_\mathbb{Y}$.

This seems to suggest that there is something peculiar about conformal prediction's (possibilistic) quantification of uncertainty, that allows it to handle the vacuous prior case more efficiently than other imprecise probability frameworks like BSA. We conjecture that the main reason for this is that CP enjoys \textit{half-coherence} \citep[Section 5.2.2]{martin2022valid}, while departing from the classical imprecise probabilistic notion of \textit{(full) coherence} \citep[Section 2.5]{walley1991statistical}. The latter entails extreme conservatism in the vacuous prior case: as we pointed out before, a vacuous prior set produces a vacuous predictive set. The gap between conformal prediction, a powerful tool that efficiently handles vacuous prior information, and the theoretically well-developed imprecise-probabilistic literature -- together with how to strike a balance between these two paradigms -- remains a largely unexplored research area, and will be the object of future work.

Another issue worth discussing is that the consonance assumption induces a ``distortion'' in the shape of the credal set $\mathcal{M}(\overline{\Pi}_\mathbf{y^n})$ induced by the conformal transducer $\pi$. With this, we mean that $\mathcal{M}(\overline{\Pi}_\mathbf{y^n})$ is ``pushed'' towards  the boundaries of the space of probabilities on $\mathbb Y$, and that the true data generating process is not its centroid. To see this, consider the following example. 

Suppose that $\mathbb{Y}=\{A,B,C\}$, and that we observe $20$ $A$'s, $30$ $B$'s and $50$ $C$'s. As a consequence of these observations, we have that vector $p^\text{emp}=(0.2,0.3,0.5)^\top$ represents the empirical pmf. Suppose that the learner chooses non-conformity measure $\psi_i(\mathbf{y^{n+1}})=1-p^\text{emp}(y_i=k)$, for all $i\in\{1,\ldots,n+1\}$. Put $T_i=\psi_i(\mathbf{y^{n+1}})$, for all $i\in\{1,\ldots,n+1\}$. We have that $\psi_i(\mathbf{y^{n+1}})=0.8$ for the $y_i$'s that are equal to $A$, $\psi_i(\mathbf{y^{n+1}})=0.7$ for the $y_i$'s that are equal to $B$, and $\psi_i(\mathbf{y^{n+1}})=0.5$ for the $y_i$'s that are equal to $C$. In turn, we derive the conformal transducer $\pi$ as
$\pi(\tilde{y}=k,\mathbf{y^n})=\frac{1}{n+1}\sum_{i=1}^{n+1} \mathbbm{1}[T_i\geq T_{n+1}]$, for all $k \in \mathbb{Y}$, so that 
$$\pi(\tilde y, \mathbf{y^n})=\begin{cases}
    \frac{21}{101} & \tilde y=A\\
    \frac{51}{101} & \tilde y=B\\
    1 & \tilde y=C
\end{cases}.$$
As we can see, our choice on non-conformity measure $\psi_i(\mathbf{y^{n+1}})$, $i\in\{1,\ldots,n+1\}$, ensures consonance. The values of the upper probabilities of the elements of $\Sigma_\mathbb{Y}=2^\mathbb{Y}$ can be found in Table \ref{table:1}. To derive them, we simply computed $\overline{\Pi}_\mathbf{y^n}(A)=\sup_{\tilde{y}\in A} \pi(\tilde{y},\mathbf{y^n})$ and $\underline{\Pi}_\mathbf{y^n}(A)=1-\overline{\Pi}_\mathbf{y^n}(A^c)$, for all $A\in 2^\mathbb{Y}$. Of course, $\overline{\Pi}_\mathbf{y^n}(\emptyset)=\underline{\Pi}_\mathbf{y^n}(\emptyset)=0$, and $\overline{\Pi}_\mathbf{y^n}(\mathbb{Y})=\underline{\Pi}_\mathbf{y^n}(\mathbb{Y})=1$.

\begin{figure}[h!]
\centering
\includegraphics[width=.3\textwidth]{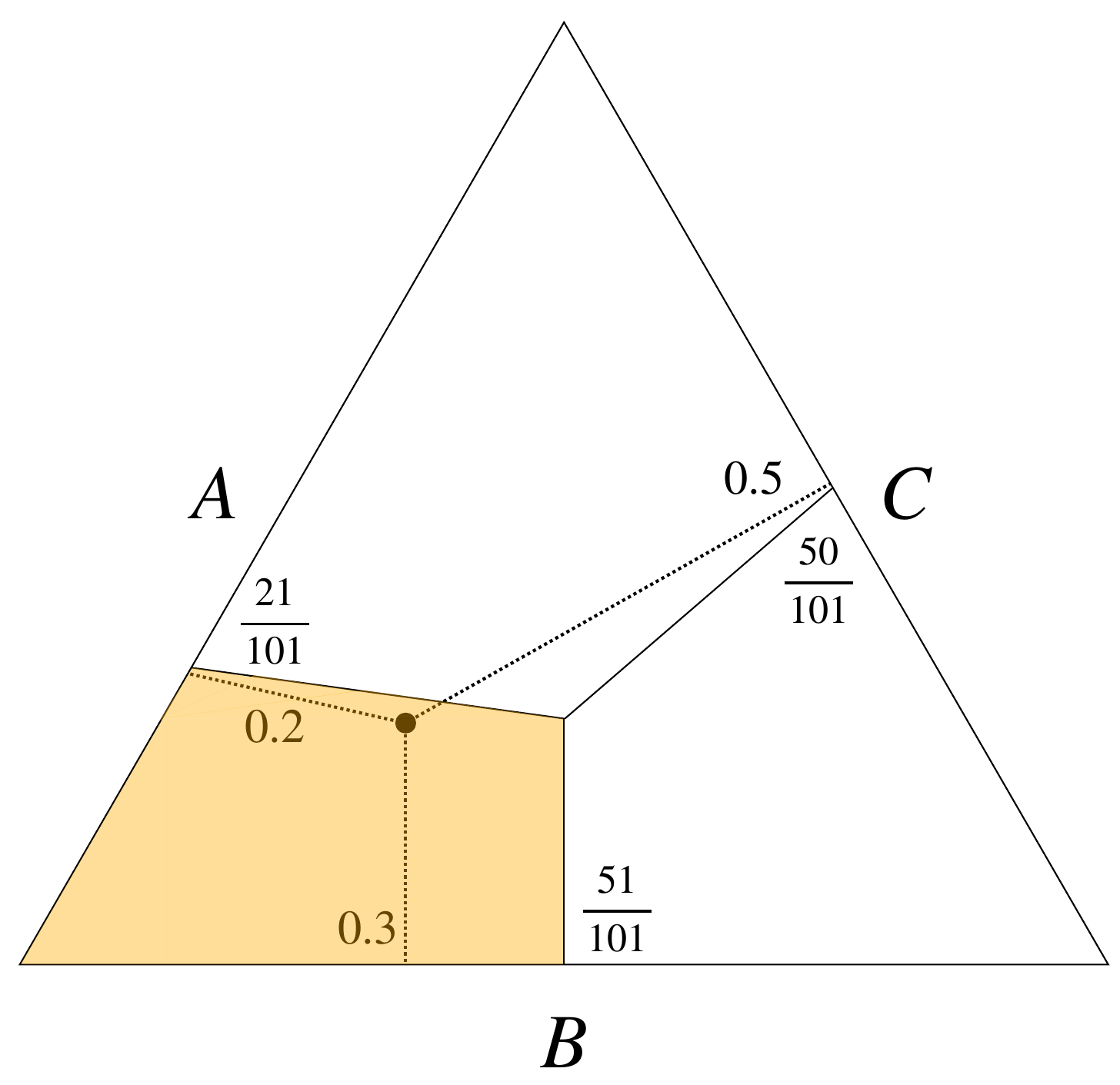}
\caption{Visual representation of $\mathcal{M}(\overline{\Pi}_\mathbf{y^n})$ in our example. As we can see, it is ``pushed'' towards the boundary of the unit simplex. We also depicted  $p^\text{emp}=(0.2,0.3,0.5)^\top$ as a black dot.}\label{fig2}
\end{figure}

\begin{table}[h]
\centering
\begin{tabular}{l|ll}
            & $\underline{\Pi}_\mathbf{y^n}$ & $\overline{\Pi}_\mathbf{y^n}$ \\ \hline
$\{A\}$     & $0$                            & ${21}/{101}$              \\ \hline
$\{B\}$     & $0$                            & ${51}/{101}$              \\ \hline
$\{C\}$     & ${50}/{101}$               & $1$                           \\ \hline
$\{A,B\}$   & $0$                            & ${51}/{101}$              \\ \hline
$\{B,C\}$   & ${80}/{101}$               & $1$                           \\ \hline
$\{A,C\}$   & ${50}/{101}$               & $1$                                                
\end{tabular}
\caption{Values that the lower and upper probabilities assign to the elements of $\Sigma_\mathbb{Y}=2^\mathbb{Y}$.}
\label{table:1}
\end{table}
We depict the credal set $\mathcal{M}(\overline{\Pi}_\mathbf{y^n})$ associated with the values of upper probability $\overline{\Pi}_\mathbf{y^n}$ in Figure \ref{fig2}. As we can see, the consonance assumption ``pushes'' $\mathcal{M}(\overline{\Pi}_\mathbf{y^n})$ towards the boundary of the unit simplex, and -- in this particular case -- towards the vertex associated with event $C$. Notice that this is not a mere consequence of the nonconformity measure that we chose in this example, but rather a characteristic ensuing from enforcing consonance. Whether this deformation has an effect on the size of the IHDR is an open question. What is apparent, though, is that it makes it difficult to gain insights around the true data generating process, which is not the centroid of $\mathcal{M}(\overline{\Pi}_\mathbf{y^n})$ \citep{miranda2023centroids}. 

In addition, if we measure the aleatoric uncertainty $\text{AU}[\mathcal{M}(\overline{\Pi}_\mathbf{y^n})]$ -- the type of uncertainty that is inherent to the data generating process, and as such, {irreducible} -- associated with $\mathcal{M}(\overline{\Pi}_\mathbf{y^n})$ via the lower entropy $\underline{H}(P)=\inf_{P \in \mathcal{M}(\overline{\Pi}_\mathbf{y^n})} H(P)$ \citep{abellan}, 
we have that $\text{AU}[\mathcal{M}(\overline{\Pi}_\mathbf{y^n})]=\underline{H}(P)=0$, because $\delta_{\{C\}} \in \mathcal{M}(\overline{\Pi}_\mathbf{y^n})$.\footnote{Here $\delta_\cdot$ denotes the Dirac measure.} While this is a feature of the measure chosen for the AU, it still seems to highlight a shortcoming of the consonance assumption.

Finally, we point out how proving or disproving a fundamental incompatibility between consonance and minimality (i.e. whether consonance might prevent the credal set from being minimal in size) is a highly non-trivial question, which we plan to inspect in future research. 

\section{Conclusion}\label{concl}
In the present work we study the conformal construction of credal sets. We show that the IHDR generated by the credal set induced by the conformal transducer $\pi$ is equivalent to the classical CPR, and it retains the same uniform probabilistic guarantee. 
Our results are devised in the general effort of framing conformal prediction as a model-free imprecise probabilistic method. This continues the undertaking of \citet{cella2022validity}, who show that CP is related to credal sets and inferential models \citep{martin2015inferential}; we prove that it is also linked to clouds. 

Notice that throughout the paper we tacitly operated under the \textit{closed world assumption} \citep{caprio2023constriction}, that is, we assumed that the support of the true data generating process does not change, and it is equal to $\mathbb{Y}$. In the future, we plan to forego this requirement. 

We also intend to extend the conformal construction of credal sets to the generalized CP frameworks studied in \citet{barber2023conformal,gibbs2023conformal}. 

Furthermore, we plan to 
(i) study in greater detail the empirical implications of our results, and (ii) inspect whether set-function $CP$ in \eqref{cp-function} is continuous (in some sense) in $\Psi$. If that were true, it would be easier to find $\Psi^\star$ that minimizes $\int_{\mathscr{R}_\alpha^\Psi(\mathbf{y^n})} \text{d}y$, that is, the non-conformity measure $\Psi^\star$ that induces the smallest possible CPR which still satisfies the $1-\alpha$ uniform probabilistic guarantee. 

In addition -- building on our considerations in Section \ref{disc} -- we plan to study the geometry of the conformally-built credal set $\mathcal{M}(\overline{\Pi}_\mathbf{y^n})$, e.g. in the spirit of \citet{cuzzolin2020geometry}. We are particularly interested in (a) its position within the simplex, and (b) the distance between the true data generating process $\mathbf{P}$ and any notion of the centroid of $\mathcal{M}(\overline{\Pi}_\mathbf{y^n})$ proposed by \citet{miranda2023centroids}.

Finally, we would like to further our investigation on the shortcomings of the consonance assumption. This is because \citep{caprio-conformalized} showed that there is at least a case of conformalized credal sets for classification problems in which such an assumption can be foregone.

\section*{Acknowledgments}
Michele Caprio was partially funded by the Army Research Office, grant number  ARO MURI W911NF2010080. He wishes to express his gratitude to Ryan Martin and Marianne Johnson for insightful discussions. Yusuf Sale is supported by the DAAD program Konrad Zuse Schools of Excellence in Artificial Intelligence, sponsored by the Federal Ministry of Education and Research. All the authors wish to express their gratitude to four anonymous reviewers for their insightful comments.

\bibliographystyle{apalike}
\bibliography{references}
\end{document}